\documentclass[twocolumn,final,authoryear]{elsarticle}
\usepackage{framed,multirow}
 
\usepackage{amssymb}
\usepackage{latexsym}

\usepackage{url}
\usepackage{xcolor}
\definecolor{newcolor}{rgb}{.8,.349,.1}

\usepackage{times}
\usepackage{epsfig}
\usepackage{amssymb}
\usepackage{booktabs}
\usepackage{makecell}

\usepackage{cancel}

\begin{document}

\begin{frontmatter}
	
\title{Top-Down Saliency Detection Driven by Visual Classification}

\author[1]{Francesca Murabito\corref{cor1}} 
\ead{fmurabit@dieei.unict.it}
\author[1]{Concetto Spampinato}
\author[1]{Simone Palazzo}
\author[1]{Daniela Giordano}
\author[2]{\\Konstantin Pogorelov}
\author[2]{Michael Riegler}
\address[1]{University of Catania, Viale Andrea Doria 6, Catania 95125, Italy}
\address[2]{Simula Research Lab, University of Oslo, Martin Linges vei 25, Fornebu and 1364, Norway}

%
%

%
%
%
\begin{abstract}
This paper presents an approach for saliency detection able to emulate the integration 
of the top-down (task-controlled) and bottom-up (sensory information) processes involved in human visual attention. In particular, we first learn how to generate saliency when a specific visual task has to be accomplished. Afterwards, we investigate if and to what extent the learned saliency maps can support visual classification in nontrivial cases. 
To achieve this, we propose \textit{SalClassNet}, a CNN framework consisting of two networks jointly trained: a) the first one computing top-down saliency maps from input images, and b) the second one exploiting the computed saliency maps for visual classification.\\ 
To test our approach, we collected a dataset of eye-gaze maps, using a Tobii T60 eye tracker, by asking several subjects to look at images from the Stanford Dogs dataset, with the objective of distinguishing dog breeds. 

Performance analysis on our dataset and other saliency benchmarking datasets, such as POET, showed that SalClassNet outperforms state-of-the-art saliency detectors, such as SalNet and SALICON. 
Finally, we also analyzed the performance of SalClassNet in a fine-grained recognition task and found out that it yields enhanced classification accuracy compared to Inception and VGG-19 classifiers.
The achieved results, thus, demonstrate that 1) conditioning saliency detectors with object classes reaches state-of-the-art performance, and 2) explicitly providing top-down saliency maps to visual classifiers enhances accuracy.
\end{abstract}

%
%
\end{frontmatter}



\section{Introduction}
\begin{figure}[t!]
	\includegraphics[width=0.234\textwidth]{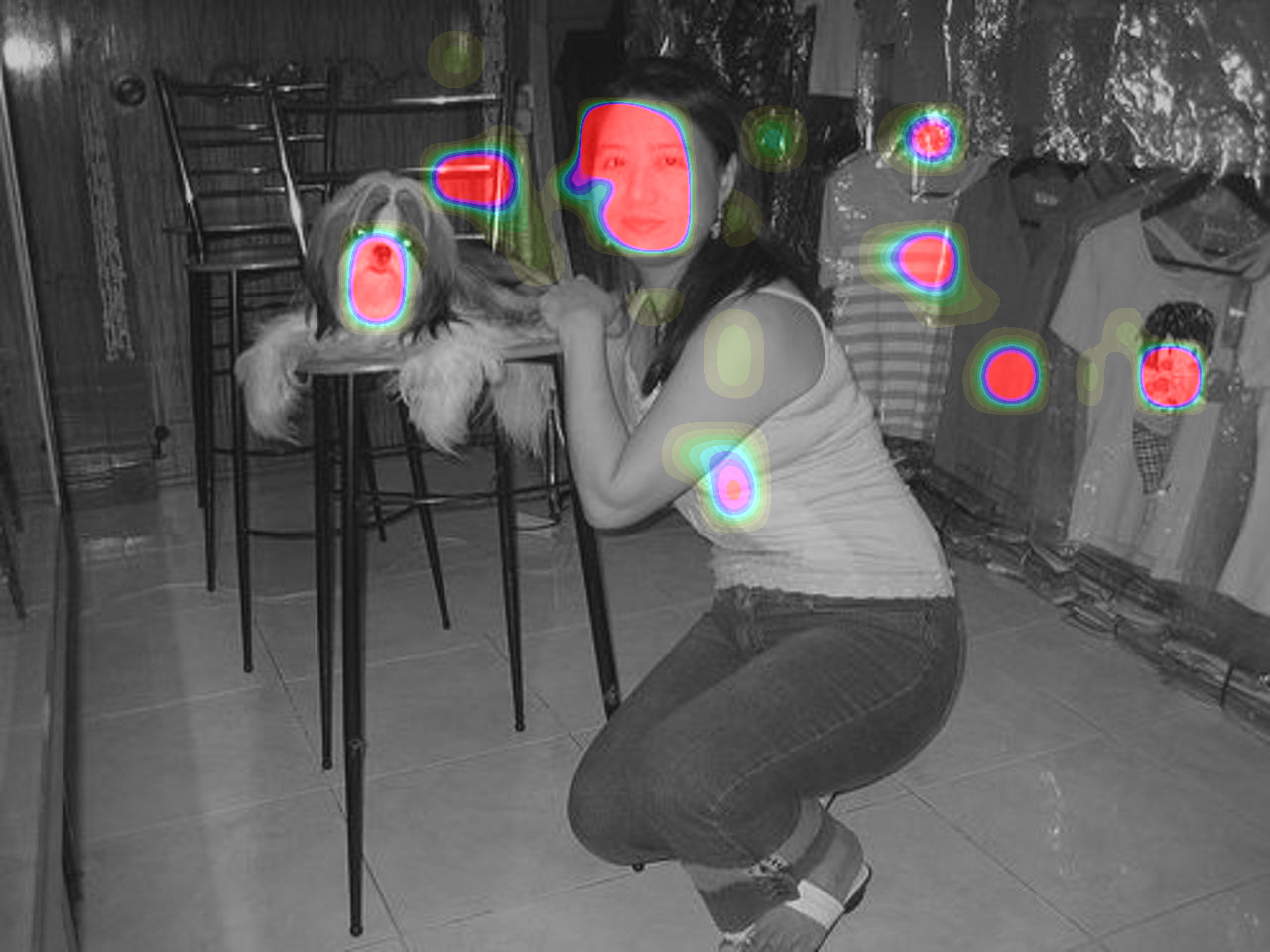}
	\includegraphics[width=0.234\textwidth]{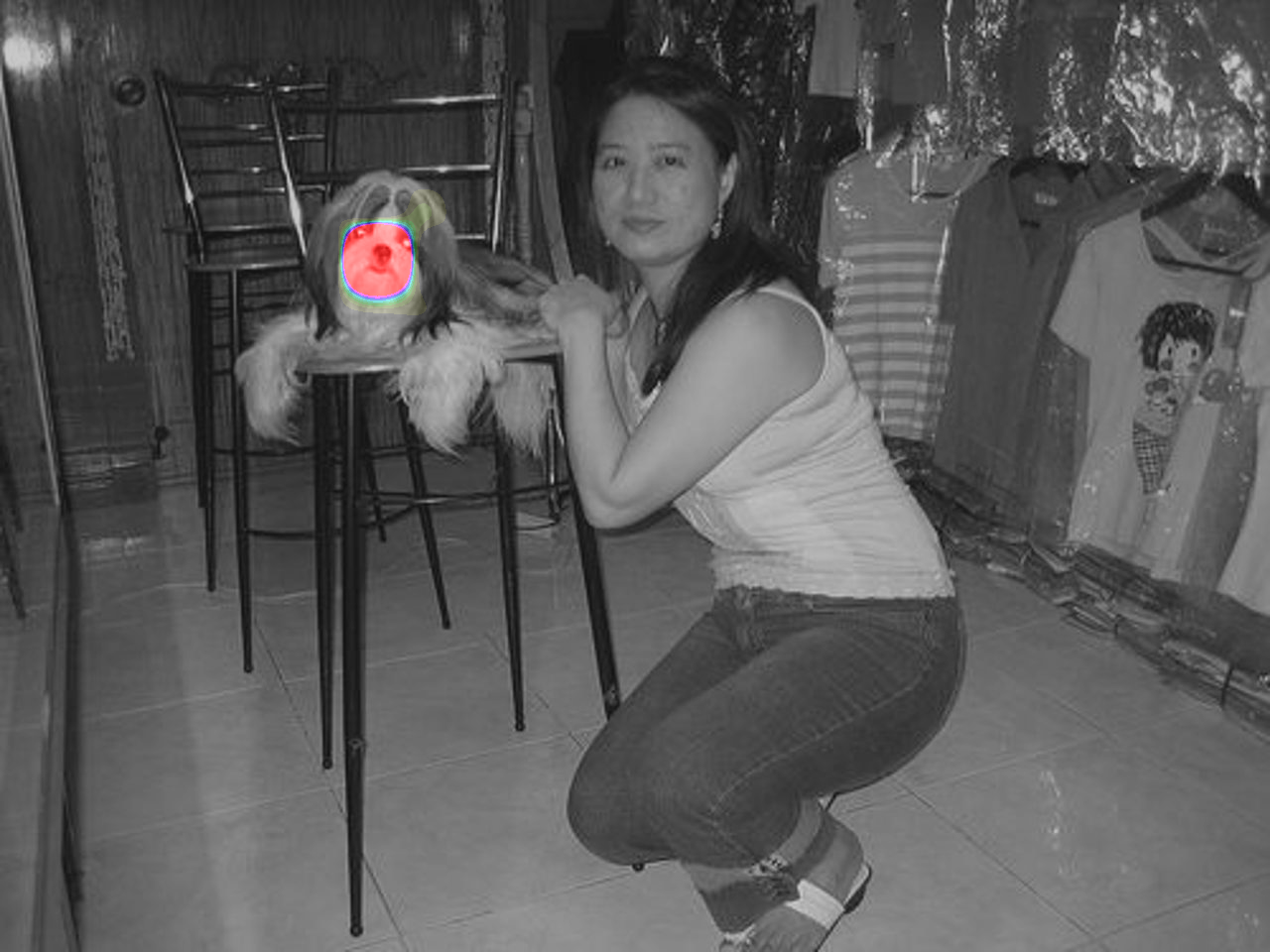}\\ 
	\\
	\includegraphics[width=0.234\textwidth]{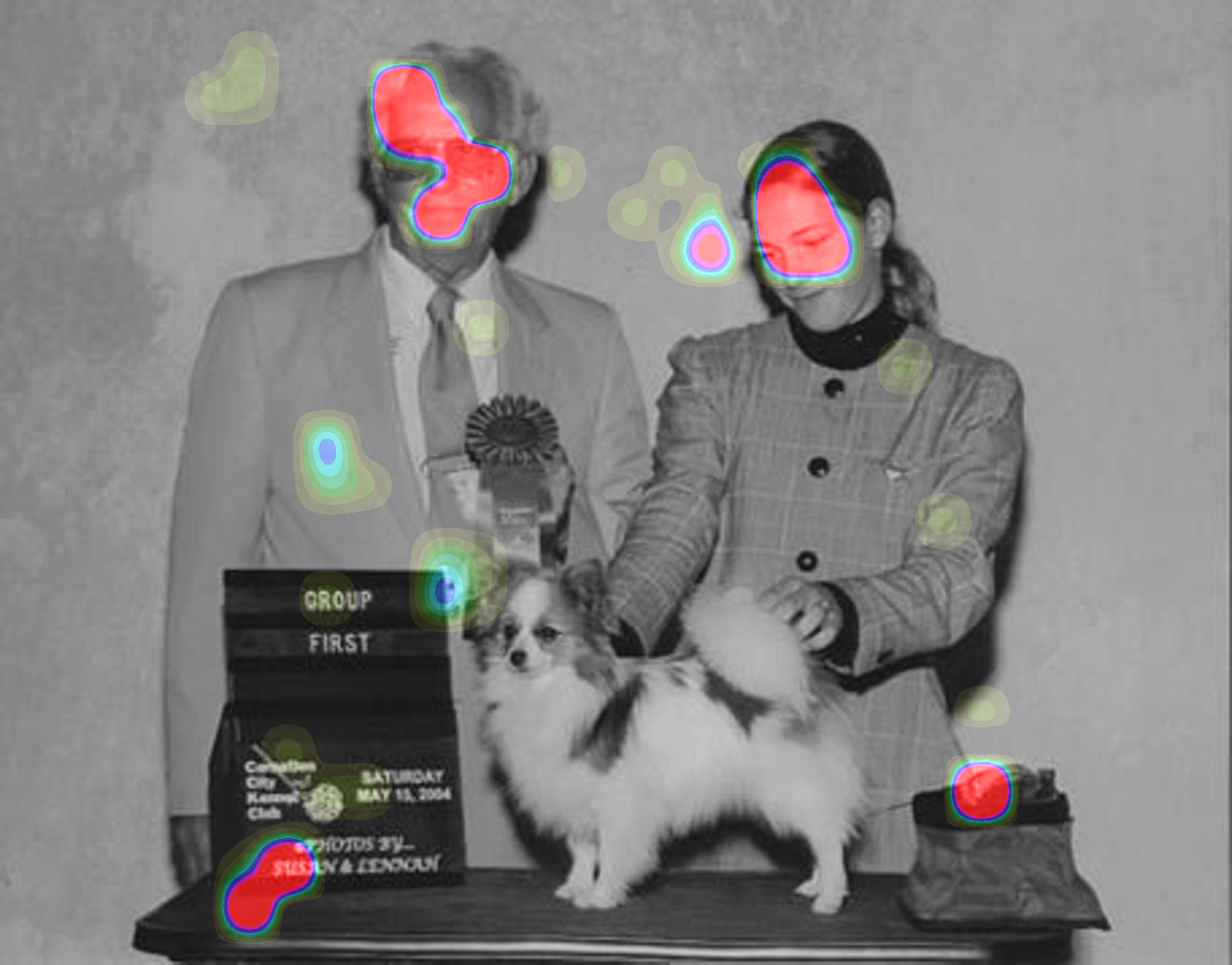}
	\includegraphics[width=0.234\textwidth]{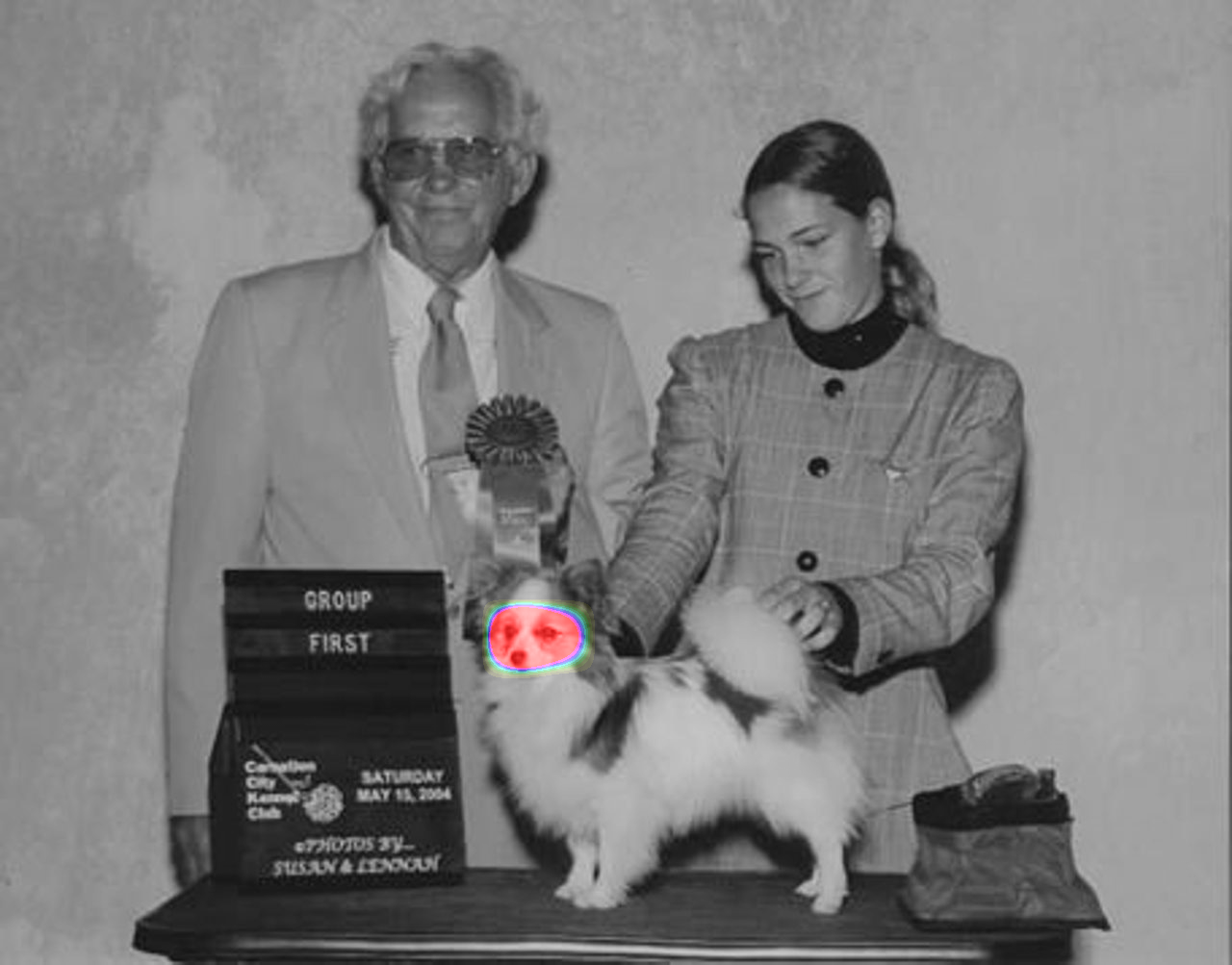}\\ 
	\\
	\includegraphics[width=0.234\textwidth]{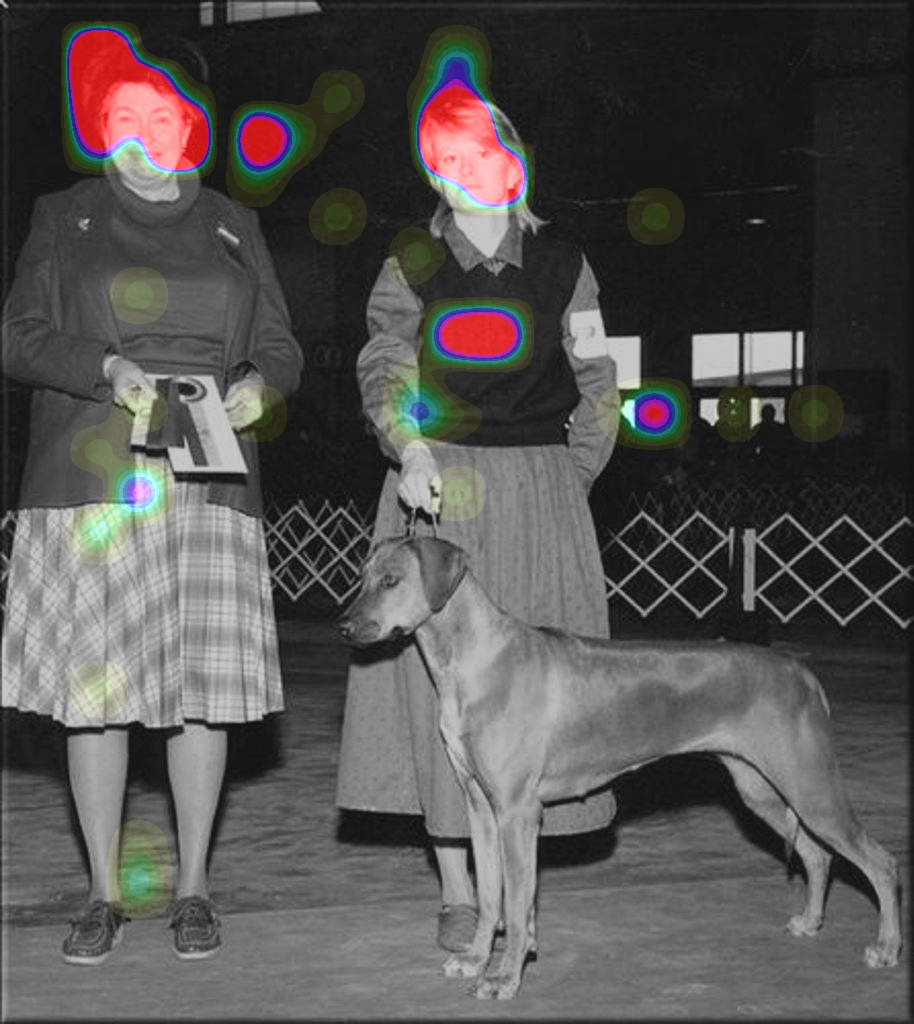}
	\includegraphics[width=0.234\textwidth]{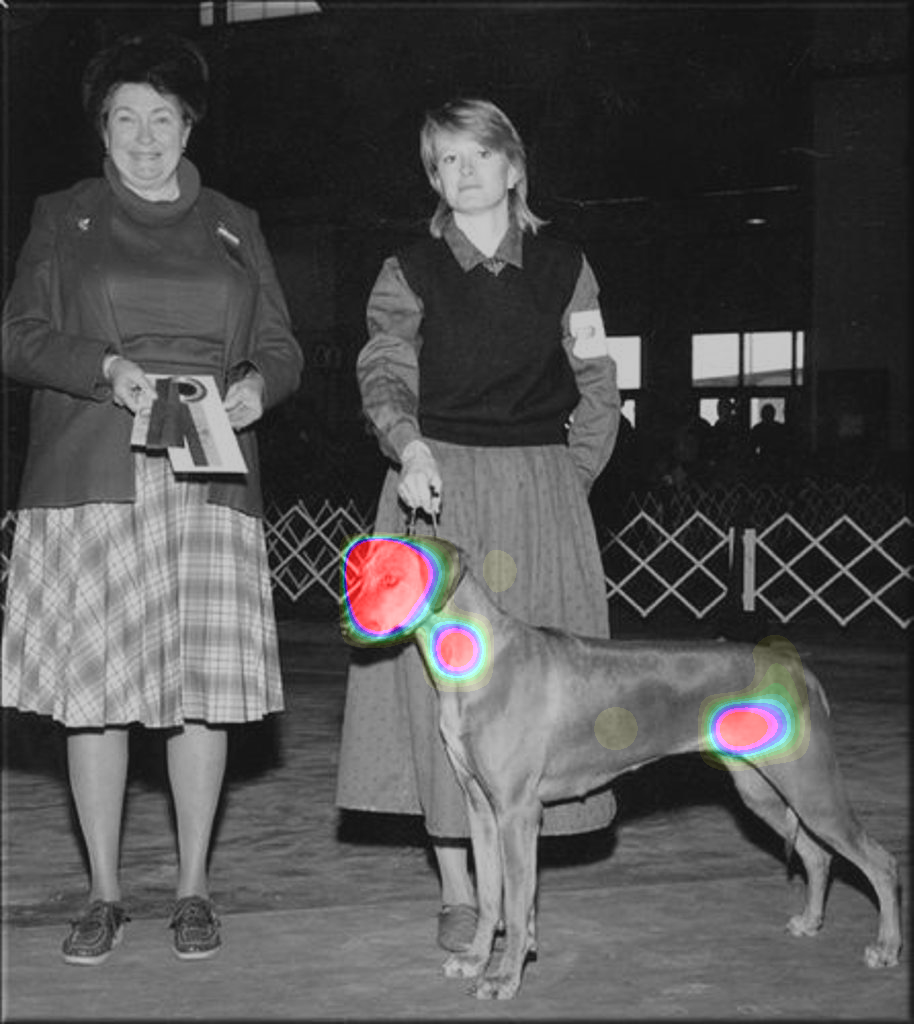}
	\caption{{\bf First column} --- Eye fixations in free-viewing experiments in images with multiple objects. Some of the salient regions cannot be used for dog species classification. {\bf Second Column} --- Eye fixation shifts when asking to guess dog breeds.}
	\label{fig:heatmaps}
\end{figure}

Computer vision and machine learning methods have long attempted to emulate humans while performing visual tasks. Despite the high intentions, the majority of the existing automated methods rely on a common schema, i.e., learning low- and mid-level visual features for a given task, often without taking into account the peculiarities of the task itself. One of the most relevant example of task-driven human process is visual attention, i.e., gating visual information to be processed by the brain according to the intrinsic visual characteristics of scenes (bottom-up process) and to the task to be performed (top-down process). Saliency detection building only on the bottom-up process mainly employs low-level visual cues, modeling unconscious vision mechanisms, and shows huge limitations in task-oriented computer vision methods. For example, traditional saliency methods~(\citealp{Itti00}) miss objects of interest in highly cluttered backgrounds since they detect visual stimuli, which often are unrelated to the task to be accomplished, as shown in Fig.~\ref{fig:heatmaps}. Analogously, image classifiers fail in cases of cluttered images as they tend to extract low and mid-level visual descriptors and match them with learned data distributions without focusing on the most salient image parts. 

Under this scenario, the contribution of this paper is twofold: a) we present a method for saliency detection guided by a classification task; and b) we demonstrate that exploiting task-based saliency maps improves classification performance. 
More specifically, we propose and train, in an end-to-end fashion, a convolutional neural network --- \textit{SalClassNet} --- consisting of two parts: the first one generating top-down (classification-guided) saliency maps from input images, while the second one taking images and the learned maps as input to perform visual categorization. 

We tested the saliency detector of SalClassNet over saliency benchmarks, where it significantly outperformed existing methods such as SalNet~(\citealp{Pan16}) and SALICON~(\citealp{Salicon}). In particular, we demonstrate how the propagation of a mixed saliency/classification loss throughout the upstream SalClassNet saliency detector is the key to learn task-guided saliency maps able to better detect the most discriminative features in the categorization process.\\

As for evaluating the performance of SalClassNet for visual categorization, we tested it on fine-grained classification tasks over the Stanford Dogs~(\citealp{KhoslaYaoJayadevaprakashFeiFei_FGVC2011}), the CUB-200-2011~(\citealp{cub}), and the Oxford Flower 102~(\citealp{flower}) datasets, showing that explicitly providing visual classifiers with saliency leads to improved performance. \\

As an additional contribution, we release our saliency dataset containing of about 10,000 maps recorded from multiple users when performing visual classification on the 120 Stanford Dogs classes, as well as with the \textit{SalClassNet} Torch code and all trained models.

\section{Related work} 
Visual attention in humans can be seen as the integration between a) an early bottom-up unconscious process where the attention is principally guided by some coarse visual stimuli, which can be local (e.g., center-surround mechanisms) or global (dependent from the context); and b) a late top-down process, which biases the observation towards those regions that consciously attract users' attention according to a specific visual task.
While the former has been extensively researched in the computer vision field with a significant number of proposed saliency detection methods~(\citealp{Li16,DeepGaze,Salicon,Pan16,superCNN,Liu15,Li15,Zhao15,Han16}), top-down processes have received much less attention (\citealp{4270335,5459462,Itti2012,ZHU201440}), mainly because of the greater difficulty to emulate high-level cognitive processes than low-level cues based on orientation, intensity and color~(\citealp{Itti00}). However, understanding the processes which are behind task-controlled visual attention may be of crucial importance to make machines see and understand the visual world as humans do and to solve complex vision tasks, such as recognition of multiple objects in cluttered scenes (\citealp{WALTHER200541}).\\
Recently, the rediscovery of convolutional neural networks and their high performance on visual tasks have led to the development of deep saliency detection networks that either adopt multi-scale patches for global/coarser and local/finer features extraction for further saliency assessment~(\citealp{superCNN,Liu15,Zhao15,Li15,Lin14,Han16,Li16, Salicon,Liu15,Shen14,Wang15CVPR,Zhao15,Tang16ECCV,DISC}) or learn, in an end-to-end fashion, saliency maps as in~\cite{Salicon, Pan16, Li16CVPR}. In particular, the recent work by \cite{Pan16} presents a fully-convolutional CNN (partly trained from scratch and partly re-using low-level layers from existing models) for saliency prediction; another fully-convolutional architecture is the one presented in~\cite{Salicon}, which processes images at two different scales and is based on deep neural networks trained for object recognition; the latter was used as basis for our work as described later.\\ 
Lately, the idea of using saliency for improving classification performance has gained significant attention from the computer vision community, coming up with saliency detection models that have been integrated into visual classification methods. In~\cite{Ren15}, saliency maps are employed to weigh features both in the learning and in the representation steps of a sparse coding process, whereas in~\cite{Zhang16} CNN-based part detections are encoded via Fisher Vectors and the importance of each descriptor is assigned through a saliency map.
\cite{Ba15} extended the recurrent attention model (RAM) presented in~\cite{Mnih14} (a model based on a combination between recurrent neural networks and reinforcement learning to identify \textit{glimpse} locations) by training it to detect and classify objects after identifying a fixed number of glimpses. 
Similarly, recent saliency detection methods have been fed with high level information in order to include top-down attention processes. In~\cite{cao2015}, given the class label as prior, the parameters of a new feedback layer are learned to optimize the target neuron output by filtering out noisy signals; in~\cite{zhang2016} a new backpropagation scheme, ``Excitation Backprop'', based on a probabilistic version of the Winner-Take-All principle, is introduced to identify task-relevant neurons for weakly-supervised localization. Our saliency maps differ from the ones computed by those methods since the only top-down signal introduced in our training is a class-agnostic classification loss; hence, our maps are able to highlight those areas which are relevant for classifying generic images.
A work similar in the spirit to ours is~\cite{Almahairi16}, where a low-capacity network initially scans the input image to locate salient regions using a gradient entropy with respect to feature vectors; then, a high-capacity network is applied to the most salient regions and, finally, the two networks are combined through their top layers in order to classify the input image. Our objective, however, is to perform end-to-end training, so that the classification error gradient can directly affect the saliency generation process.
Given these premises, the most interesting saliency network architectures for our purpose are the fully-convolutional ones, whose output can be seamlessly integrated into a larger framework with a cascading classification module. Tab.~\ref{tab:method_comparison} summarizes the results of state-of-the-art fully-convolutional saliency networks on a set of commonly-employed datasets for saliency detection benchmarking, namely SALICON validation and test sets~(\citealp{Salicon}), iSUN validation and test sets~(\citealp{isun}) and MIT300~(\citealp{mit300}). In this work, we focus our attention, both as building blocks and evaluation baselines, on the SALICON~(\citealp{Salicon}) and SalNet~(\citealp{Pan16}) models, thanks to code availability and their fully-convolutional nature.

 \begin{table*}[htb]
	\centering
	\resizebox{\textwidth}{!}{
		\begin{tabular}{cccccccccc}
			\toprule
			\textbf{Method} & \textbf{N. Layers} & \textbf{Framework}  & \textbf{Training Dataset} & \textbf{SALICON Test} & \textbf{SALICON Val} & \textbf{iSUN Test} & \textbf{iSUN Val} & \textbf{MIT300} \\
			\midrule
			\textbf{JuntingNet} & 5 & Lasagne & SALICON  & \makecell[l]{CC = 0.60 \\ Shuffled-AUC = 0.67 \\ AUC Borji = 0.83} & \makecell[l]{CC = 0.58 \\ Shuffled-AUC = 0.67 \\ AUC Borji = 0.83} & \makecell[l]{CC = 0.82 \\ Shuffled-AUC = 0.67 \\ AUC Borji = 0.85} & \makecell[l]{CC = 0.59 \\ Shuffled-AUC = 0.64 \\ AUC Borji = 0.79} &  \makecell[l]{CC = 0.53 \\ Shuffled-AUC = 0.64 \\ AUC Borji = 0.78} \\
			\hline
			\textbf{SalNet} & 10 & Caffe & SALICON & \makecell[l]{CC = 0.62 \\ Shuffled-AUC = 0.72 \\ AUC Borji = 0.86} & \makecell[l]{CC = 0.61 \\ Shuffled-AUC = 0.73 \\ AUC Borji = 0.86} & \makecell[l]{CC = 0.62 \\ Shuffled-AUC = 0.72 \\ AUC Borji = 0.86} &  \makecell[l]{CC = 0.53 \\ Shuffled-AUC = 0.63 \\ AUC Borji = 0.80} & \makecell[l]{CC = 0.58 \\ Shuffled-AUC = 0.69 \\ AUC Borji = 0.82} \\
			\hline
			\textbf{SALICON} & 16 & Caffe & OSIE  & --- & --- & --- & ---  & \makecell[l]{CC = 0.74 \\ Shuffled-AUC = 0.74 \\ AUC Borji = 0.85} \\
			\hline
			\textbf{DeepGaze} & 5 & Not Available & MIT1003  &--- & --- & --- & ---  & \makecell[l]{CC = 0.48 \\ Shuffled-AUC = 0.66 \\ AUC Borji = 0.83 } \\
			\hline
			\textbf{DeepGaze 2} & 19 & Web Service & SALICON - MIT1003  & --- & --- & --- & \makecell[l]{CC = 0.51 \\ Shuffled-AUC = 0.77 \\ AUC Borji = 0.86} \\
			\hline
			\textbf{ML-NET} & 19 + 2  & Theano & SALICON  & \makecell[l]{CC = 0.76 \\ Shuffled-AUC = 0.78} & --- & --- & --- &  \makecell[l]{CC = 0.69 \\ Shuffled-AUC = 0.70 \\ AUC Borji = 0.77} \\
			\hline
			\textbf{DeepFix} & 20 & Not Available & SALICON & --- & --- & --- & --- & \makecell[l]{CC = 0.78 \\ Shuffled-AUC = 0.71 \\ AUC Borji = 0.80} \\
			\hline 
			\textbf{eDN} & Ensemble & Sthor & MIT1003  &--- & --- & --- & --- & \makecell[l]{CC = 0.45 \\ Shuffled-AUC = 0.62 \\ AUC Borji = 0.81} \\
			\hline  
			\textbf{PDP} & 16 + 3 & Not Available & SALICON  & \makecell[l]{CC = 0.77 \\ Shuffled-AUC = 0.78  \\ AUC Borji = 0.88} & \makecell[l]{CC = 0.74 \\ Shuffled-AUC = 0.78} & --- & --- & \makecell[l]{CC = 0.70 \\ Shuffled-AUC = 0.73 \\ AUC Borji = 0.80} \\
			\bottomrule
	\end{tabular}}
	\caption{A summary of state-of-art fully-convolutional methods and their results, according to the most common metrics, on several saliency datasets. Dataset references: \textbf{SALICON Test and Val}: \cite{salicon_dataset}; \textbf{iSUN Test and Val}: \cite{isun}; \textbf{MIT300}: \cite{mit300}. Method references: \textbf{JuntingNet and SalNet}: \cite{Pan16}; \textbf{SALICON}: \cite{Salicon}; \textbf{DeepGaze}: \cite{DeepGaze}; \textbf{DeepGaze2}: \cite{deepgaze2}; \textbf{ML-NET}: \cite{cornia2016multi}: \textbf{DeepFix}: \cite{deepfix}; \textbf{eDN}: \cite{vig2014large};  \textbf{PDP}: \cite{jetley2016end}.}.
	\label{tab:method_comparison}
\end{table*}

\section{SalClassNet: A CNN model for top-down saliency detection}
\label{sec:model}
\begin{figure*}[htb]
	\centering
	\includegraphics[width=1\textwidth]{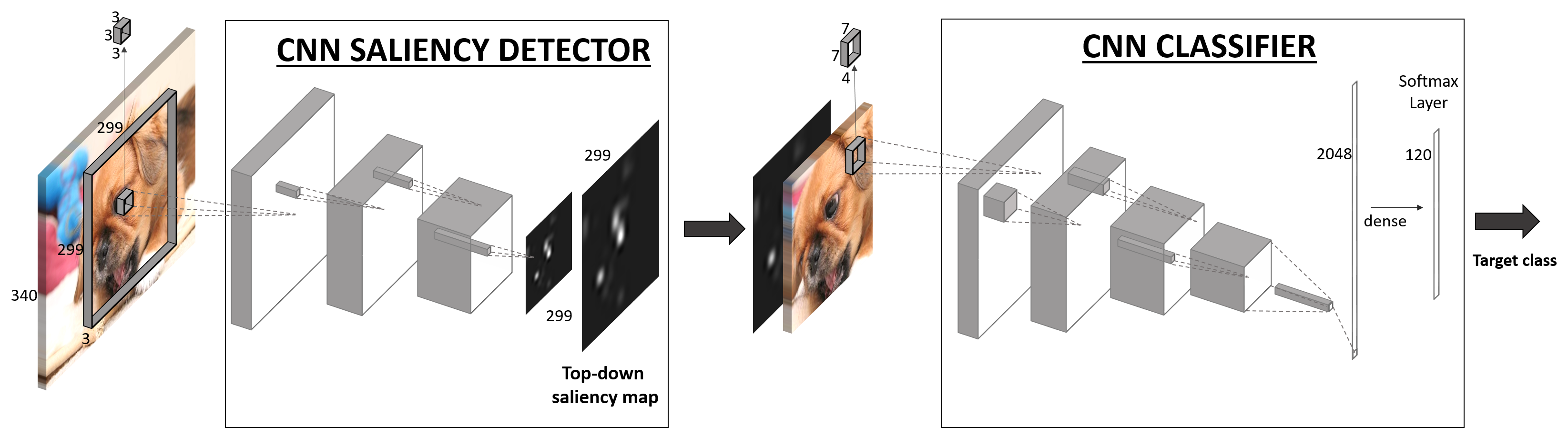}
	\caption{Architecture of the proposed model -- \textit{SalClassNet}-- for saliency detection guided by a visual classification task. Input images are processed by a saliency detector, whose output together with input images are fed to a classification network with 4-channel first-layer kernels for processing image color and saliency and providing image classes as output.}
	\label{fig:salclassnet}
\end{figure*}

The general architecture of our network is shown in Fig.~\ref{fig:salclassnet} and is made up of two cascaded modules: a saliency detector and a visual classifier, which are jointly trained in a multi-loss framework.

\subsection{Top-down saliency detection network}
\vspace{2pt} 
Although we will discuss the details of the employed saliency dataset and its generation process in Sect.~\ref{sec:dataset}, it is necessary to introduce some related information at this stage, which is important to understand the overall model.

In the dataset generation protocol, human subjects were explicitly asked to look at images and to guess their visual classes (e.g., dog breeds). Therefore, our experiments aimed to enforce top-down saliency driven by a specific classification task, rather than bottom-up saliency. In other words, instead of emphasizing the location of image regions which are visually interesting per se (which, of course, may include the target object), our visual attention maps focus on the location of features needed for identifying the target classes, ignoring anything else that may be salient but not relevant to the classification task. Hence, our saliency detector has to be able, given an input image, to produce a map of the most salient image locations useful for classification purposes. \\
To accomplish that, we propose a CNN-based saliency detector composed by thirteen  convolutional and five max pooling layers taken from VGG-19~(\citealp{Simonyan14c}). The output of the last pooling layer, i.e., 512$\times$10$\times$10 feature maps (for a 3$\times$299$\times$299 input image), is then processed by a 1$\times$1 convolution to compute a saliency score for each ``pixel'' in the feature maps of the previous layer, producing a single-channel map. Finally, in order to generate the input for the subsequent classification network, the 10$\times$10 saliency maps are upsampled to  $299\times 299$ (which is the default input size of the next classification module)  through bilinear interpolation.

As for the size of the output maps, it has to be noted that saliency is a primitive mechanism, employed by humans to drive the attention towards objects of interest, which is evoked by coarse visual stimuli (\citealp{Itti00}). Thus, increasing the resolution of saliency maps for identifying finer image details from a visual scene is not necessary, beside introducing noisy information potentially affecting negatively the classification performance (indeed, when we increased the saliency map size, the saliency accuracy did not improve). Therefore, in spite of the low spatial resolution of saliency maps, our experiments (see Sect.~\ref{sec:results}) show that the 10$\times$10 feature maps are able to encode the information needed to detect salient areas and to drive a classifier with them.

\subsection{Saliency-based classification network} 
\vspace{3pt} 
Our visual classifier is a convolutional neural network which receives as input a 4-channel RGBS image, combining the RGB image with the corresponding saliency (S) map, and provides as output the corresponding class. The underlying idea is that the network should employ those salient regions (as indicated by the input saliency map S) which are more meaningful for classification purposes.

This network is based on the Inception network~(\citealp{Szegedy15}), which comprises sixteen convolutional and one fully connected layer followed by a final softmax layer, with the first-layer convolutional kernels modified to support the 4-channel input. In particular, the 32 3$\times$3$\times$3 kernels in the first layer are converted into 32 4$\times$3$\times$3 kernels, whose weights corresponding to the RGB channels are taken from a pre-trained version of Inception network (see next Sect.~\ref{sec:results}), whereas the new weights, corresponding to the saliency input, are randomly initialized.  
Since the model includes a combination of trained weights (the ones from the original Inception) and untrained weights (the ones related to the saliency channel) we set different learning rates in order to speed up the convergence of untrained weights while not destabilizing the already learned ones.

\subsection{Multi-loss saliency-classification training} 
\vspace{3pt} 
The networks described in the previous sections are joined together into a single sequential model and trained using RGB images as input and the corresponding class labels as output. We introduced a batch normalization module between the saliency detector and the classifier, to enforce a zero-mean and unitary--standard-deviation distribution at the classifier's input. During training,
we minimize a multi-loss objective function given by a linear combination of cross-entropy classification loss $\mathcal{L_C}$, and saliency detection loss $\mathcal{L_S}$ computed as the mean square error (MSE) of the intermediate saliency detector's output (obtained after the last upsampling layer) with respect to the ground-truth saliency map for the corresponding input image:
\begin{equation}
 \mathcal{L}\left(\textbf{y}, \textbf{Y}, t, \textbf{T}\right) =  \alpha~\mathcal{L}_C\left(\textbf{y}, t\right) + \mathcal{L}_S\left(\textbf{Y}, \textbf{T}\right)
 \label{eq:multiloss}
\end{equation}
where
\begin{equation}
\mathcal{L}_C(\textbf{y},t) = - \sum_{i=1}^{n}\mathbb{I}(i = t) \log(y_i)
\end{equation}
\begin{equation}
\mathcal{L}_S(\textbf{Y},\textbf{T}) = \frac{1}{hw}\sum_{i=1}^{h}\sum_{j=1}^{w} (Y_{ij} - T_{ij})^2
\end{equation}

where $\mathcal{L}_C$ is the cross-entropy loss computed for the softmax output vector $\textbf{y}$ and the correct class $t$, $n$ indicates the number of classes in the dataset, $\mathcal{L}_S$ is the mean square error loss computed on the saliency detector's output map $\textbf{Y}$ and the ground-truth heatmap $\textbf{T}$, $h$ and $w$ are the size of the heatmap, and $\mathbb{I}(p)$ is the indicator function, which returns $1$ if $p$ is true; bold symbols denote vectors (lower case) and matrices (upper case).

The adopted multi-loss affects the model in several ways. First of all, backpropagating the classification loss to the saliency detector forces it to learn saliency features useful for classification. Secondly, backpropagating the mean square error on the saliency maps ensures that the saliency detector does not degenerate into identifying generic image features and become a convolutional layer as any other. 

\begin{figure}[htb]
 \centering
 \includegraphics[width=0.23\textwidth]{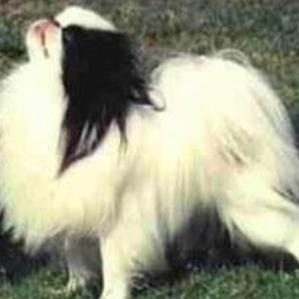}
 \includegraphics[width=0.23\textwidth]{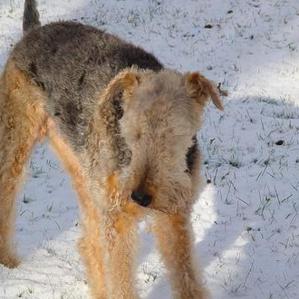}\\ \vspace{0.1cm}
 \includegraphics[width=0.23\textwidth]{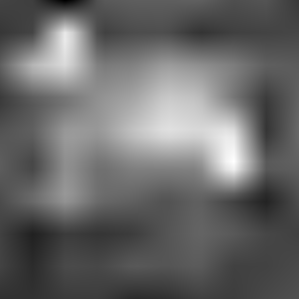}
 \includegraphics[width=0.23\textwidth]{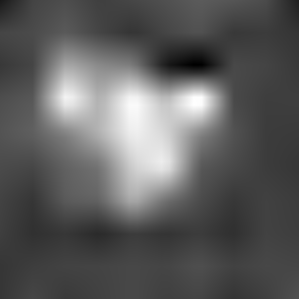}\\ \vspace{0.1cm}
 \includegraphics[width=0.23\textwidth]{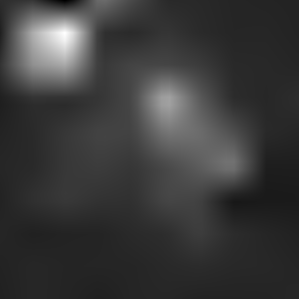}
 \includegraphics[width=0.23\textwidth]{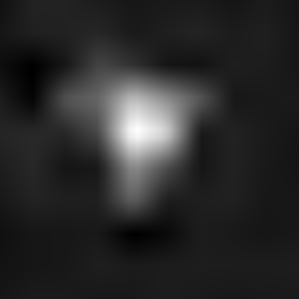}
 \caption{{\bf From saliency maps including only sensory information (bottom-up attention processes) to maps integrating task-related information (top-down processes)}. (\textit{Top row}) Two example images. (\textit{Middle row}) Bottom-up saliency maps generated by our CNN-based saliency detector fine-tuned over the Stanford Dog dataset using ground-truth heatmaps. (\textit{Bottom row}) Shift of saliency guided by the classification task, as resulting from training SalClassNet.}
 \label{fig:salnet_ft}
\end{figure}

Fig.~\ref{fig:salnet_ft} shows two output examples of how saliency changes when using only saliency loss $\mathcal{L}_S$ to train the saliency detector and when driving it by the classification loss $\mathcal{L}_C$: the saliency is shifted from generic scene elements to more discriminative features.

\section{Top-down Saliency Dataset}
\label{sec:dataset}
To test our saliency detector, we built a top-down saliency dataset -- \emph{SalDogs} -- consisting of eye-gaze data recorded from multiple human subjects while observing dog images taken from the Stanford Dogs dataset ~(\citealp{KhoslaYaoJayadevaprakashFeiFei_FGVC2011}), a collection of 20,580 images of dogs from 120 breeds (about 170 images per class). From the whole Stanford Dogs dataset, we used a subset of 9,861 images keeping the original class distribution. The eye-gaze acquisition protocol involved 12 users, who underwent breed-classification training sessions (randomly showing dog images with the related classes), and then were asked to identify the learned breeds from images. To guide top-down visual attention of participants, according to psychology research (\citealp{pmid22984991}), images were blurred with a Gaussian filter whose variance was initially set to 10 and then gradually reduced by 1 each half second until subjects were able to recognize their classes or they were completely de-blurred. Users took, on average, 2.6 seconds to identify dog breeds and 2,763 images were not identified till the end of the de-blurring process. Eye-gaze gaze were recorded through a 60-Hz Tobii T60 eye-tracker.
Tab.~\ref{tab:dataset_summary} provides an overview of the \emph{SalDogs} dataset. To the best of our knowledge, this is one of the first {\color{black}publicly-available datasets with saliency maps driven by visual classification tasks, and the first one dealing with a large number of fine-grained object classes.}

\begin{table}[htb]
	\centering
	\begin{tabular}{lc}
		\toprule
		& \textbf{Our Dataset} \\
		\midrule
		Number of images & 9,861\\
		Number of classes & 120\\
		Avg. number of images per class & 82.2\\
		Avg. number of fixation points per image & 6.2\\
		\bottomrule
	\end{tabular}
	\caption{Information on the generated saliency dataset.}
	\label{tab:dataset_summary}
\end{table}

A dataset similar to ours is POET ~(\citealp{papadopoulos2014}), which, however, does not deal with fine-grained classification tasks, but with classification at the basic level and with much fewer classes (10 Pascal VOC classes vs 120 in our case). 
Tab.~\ref{tab:dataset_comparison} reports a comparison, in terms of enforced attention mechanism (e.g., tasks accomplished by participants), number of viewers, collected images and acquisition devices, between our dataset and recent saliency benchmarking datasets. Finally, to test the generalization capabilities of our saliency detector, we also collected eye gaze data from the same 12 subjects, employing the same data acquisition protocol described above, on: a) bird images (referred in the following as \emph{SalBirds}), using a subset of 400 images taken from CUB-200-2011 dataset~(\citealp{cub}), an image dataset containing 11,788 images from 200 classes representing different bird species; and b) flower images (referred as \emph{SalFlowers}) by selecting 400 images from Oxford Flowers-102~(\citealp{flower}), which contains over 8,000 images from 102 different flower varieties.

 \begin{table*}\footnotesize
 	\centering
 	 \scriptsize{
 	\begin{tabular}{llllllll}
 		\toprule
 		Dataset & Capture method & Task &Viewers & Train & Validation &  Test & Tot \\
 		\midrule
 		SALICON~(\cite{salicon_dataset}) & Mouse clicks & Free-viewing &Crowd & 10,000 & 5,000 & 5,000 & 20,000\\
 		iSUN~(\cite{isun}) & Camera-based eye tracker & Free-viewing& Crowd & 6,000 & 926 & 2,000 & 8,926\\
 		MIT300~(\cite{mit300}) & ISCAN video-based eye tracker & Free-viewing&39 & - & - & - & 300\\
 		CAT2000~(\cite{cat2000}) & Eyelink 1000 eye tracker & Free-viewing & 24 & 2,000 & - & 2,000 & 4,000 \\
 		FIGRIM~(\cite{figrim})& Eyelink 1000 eye tracker & Memory & 15 & - & - & - &  2,787  \\
 		EyeCrowd~(\cite{eyecrowd})& Eyelink 1000 eye tracker  & Free-viewing & 16 & 450 & - & 50 & 500 \\
 		OSIE~(\cite{osie}) & Eyelink 1000 eye tracker & Free-viewing & 15 & 500 & & 200 & 700 \\
 		PASCAL-S~(\cite{pascals}) & Eyelink 1000 eye tracker & 	Free-viewing & 8 & - & - & - & 850 \\
 		ImgSal~(\cite{imgsal})& Tobii T60 eye tracker & Free-viewing & 21 & - & - & - & 235  \\
 		POET~(\cite{papadopoulos2014}) & Eyelink 2000 eye tracker & Basic  classification &28 & 441 & - & 5,829 & 6,270 \\
 		SalDogs & Tobii T60 eye tracker & Fine-grained  classification&12& 8,005 & 928 & 928\\
 		\bottomrule
 	\end{tabular}
 	\caption{Comparison between our dataset and others from the state of the art.}}
 	\label{tab:dataset_comparison}
 \end{table*}

\section{Performance analysis}
\label{sec:results}
The performance analysis focuses on assessing the quality of our model and its comparison to state-of-the-art approaches on two tasks: a) generating task-driven saliency maps from images; b) fine-grained visual recognition task.

\subsection{Datasets}
The main benchmarking dataset used for the evaluation of both saliency detection and classification models was \emph{SalDogs} (9,861 images with heatmaps), which was split into training set (80\%, 8,005 images -- \emph{SalDogs-train}), validation set (10\%, 928 images -- \emph{SalDogs-val}) and test set (10\%, 928 images -- \emph{SalDogs-test}).


Specifically for saliency detection, we also employed the POET, \emph{SalBirds} and \emph{SalFlowers} datasets (described in Sect.~\ref{sec:dataset}) to assess the generalization capabilities of the models trained on \emph{SalDogs}.


For visual classification evaluation, we first carried out a comparison of different models on \emph{SalDogs}, aimed at investigating the contribution of visual saliency to classification. Then, we assessed the generalization capabilities of SalClassNet on the CUB-200-2011 and Oxford Flower 102 fine-grained datasets.

All classification networks (SalClassNet and baseline) were first pre-trained on a de-duped version of ImageNet, obtained by removing from ImageNet the 120 classes present in the Stanford Dogs Dataset. This guarantees fairness between models regardless of pre-training: indeed, since the whole Stanford Dogs is included in ImageNet, publicly-available pre-trained VGG-19 and Inception models would have the advantage of having been trained on images included in \emph{SalDogs-test}.

\subsection{Training details}
\vspace{2pt} 
The saliency detector in SalClassNet consists of a cascade of convolutional feature extractors initialized from a pre-trained VGG-19, followed by a layer (to train from scratch) which maps each location of the final feature map into a saliency score. An initial pre-training stage was carried out on \emph{OSIE} (\citealp{osie}), as done also in SALICON. This pre-training employed mini-batch SGD optimization (learning rate: 0.00001, momentum: 0.9, weight decay: 0.0005, batch size: 16) of the MSE loss between the output and target saliency maps; data augmentation was performed by rescaling each image (and the corresponding ground-truth heatmap) to 340 pixels on the short side, while keeping aspect ratio, and randomly extracting five 299$\times$299 crops, plus the corresponding horizontal flips. After this initial pre-training, the resulting model was fine-tuned on \emph{SalDogs-train}: the learning rate was initialized to 0.001 and gradually reduced through the $1/t$ decay rule, i.e., at iteration $i$ it was computed as $l/(1 + 10^{-5}\cdot i)$, with $l$ being the initial learning rate. During this fine-tuning stage, the same data augmentation approach described above and the same values for the other hyperparameters were used.

The saliency-based classifier module of SalClassNet was initially pre-trained as a regular Inception network. Due to the inclusion of Stanford Dogs in ImageNet, we did not employ a publicly-available pre-trained network, and instead trained an Inception architecture from scratch on the de-duped version of ImageNet described in the previous section. We trained the model for 70 epochs, using mini-batch SGD for optimization, with a learning rate schedule going from 0.01 to 0.0001 over the first 53 epochs, weight decay 0.0005 up to the 30\textsuperscript{th} epoch (and 0 afterwards), momentum 0.9 and batch size 32. Data augmentation on the input images was performed as described above. 
After this pre-training was completed, we modified the first-layer  kernels to support RGB color plus saliency input, by adding a dimension with randomly-initialized weights to the relevant kernel tensors, and we fine-tuned the model on \emph{SalDogs-train} for classification, passing as input, each image with the corresponding ground-truth saliency map. Since some weights in the model had already been pre-trained and others had to be trained from scratch, the learning rate was initially set to 0.05 for the untrained parameters, and to 0.001 for the others. We used the same procedures for learning rate decay and data augmentation as in the fine-tuning of the saliency detector, and a batch size of 16.

\begin{figure}[htb]
	\centering
	\includegraphics[width=0.5\textwidth,height=4.8cm]{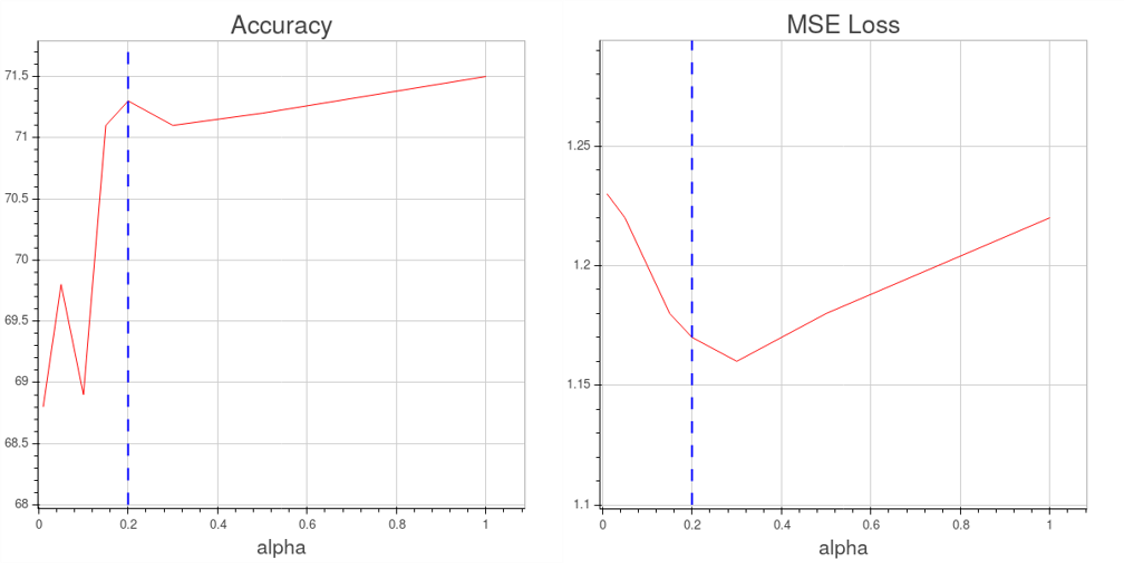}
	\caption{Classification accuracy and MSE w.r.t. $\alpha$ values: 0.2 was chosen as the best trade-off between the two performance metrics.}
	\label{fig:losses_vs_alpha}
\end{figure}

The final version of the SalClassNet model - which is the one employed in the following experiments - was obtained by concatenating the saliency detector and the saliency-based classifier and fine-tuning it, in an end-to-end fashion, on \emph{SalDogs-train}. Indeed, up to this point, the saliency detector had never been provided with an error signal related to a classification loss, as well as the saliency-based classifier had never been provided with input maps computed by an automated method. Again, the previous procedures for data augmentation and learning rate decay were employed, with a single initial learning rate of 0.001. The $\alpha$ value in Eq.~\ref{eq:multiloss}, weighing the classification loss with respect to the saliency MSE loss, was set to 0.2, since it provided the best accuracy trade-off (see Fig.~\ref{fig:losses_vs_alpha}).

During the fine-tuning stages of the individual modules and of the end-to-end model, at the end of each epoch we monitored the classification accuracy and the saliency MSE loss over \emph{SalDogs-val} (evaluating only the central crop of each rescaled image), and stopped training when both had not improved for 10 consecutive epochs: in practice, all models converged in 70-120 epochs. Model selection was performed by choosing the model for which the best relevant accuracy measure (MSE loss for the saliency detector, classification accuracy for the saliency-based classifier and the full SalClassNet model) had been obtained.

\subsection{{\color{black}Saliency detection performance}}
\begin{figure*}[h]
	\centering
	\includegraphics[width=0.13\textwidth]{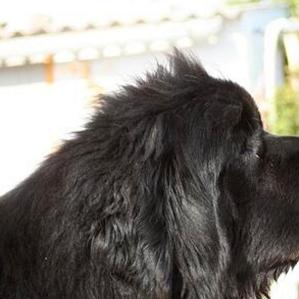}
	\includegraphics[width=0.13\textwidth]{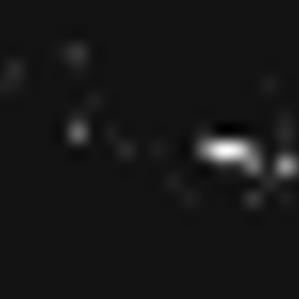}
	\includegraphics[width=0.13\textwidth]{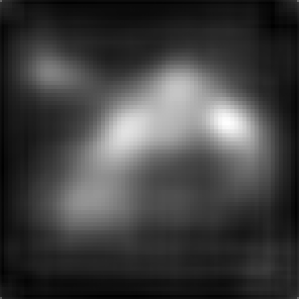}
	\includegraphics[width=0.13\textwidth]{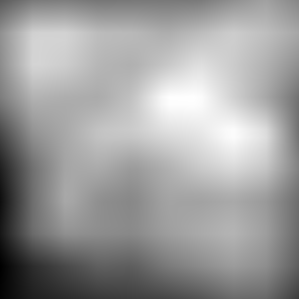}
	\includegraphics[width=0.13\textwidth]{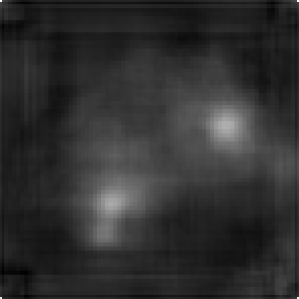}
	\includegraphics[width=0.13\textwidth]{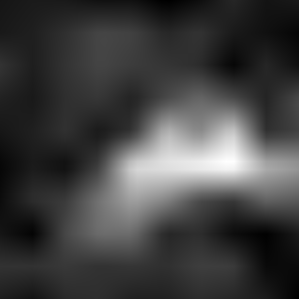}
	\includegraphics[width=0.13\textwidth]{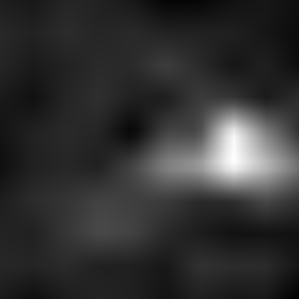}\\ 
	\vspace{0.05cm}
	\includegraphics[width=0.13\textwidth]{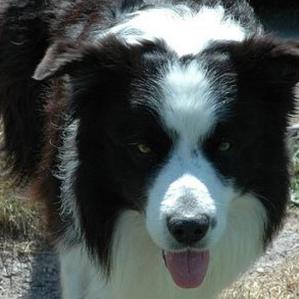}
	\includegraphics[width=0.13\textwidth]{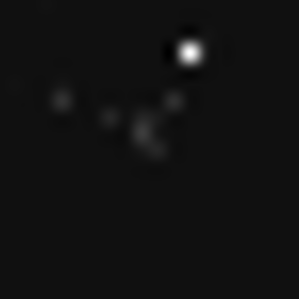}
	\includegraphics[width=0.13\textwidth]{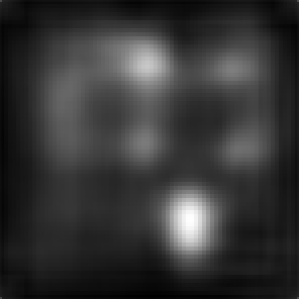}
	\includegraphics[width=0.13\textwidth]{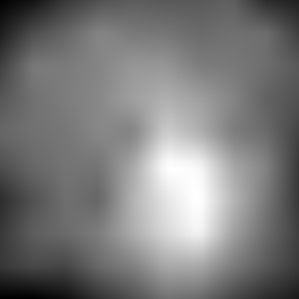}
	\includegraphics[width=0.13\textwidth]{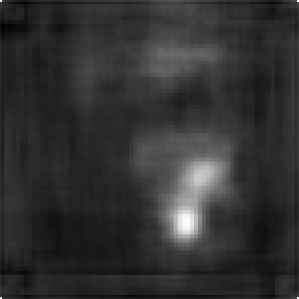}
	\includegraphics[width=0.13\textwidth]{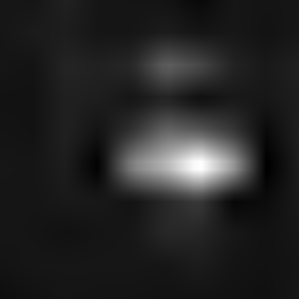}
	\includegraphics[width=0.13\textwidth]{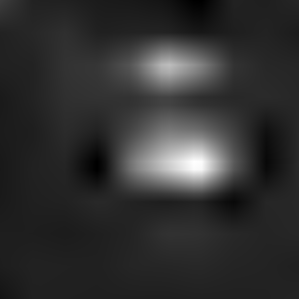}\\ 
	\vspace{0.05cm}
	\includegraphics[width=0.13\textwidth]{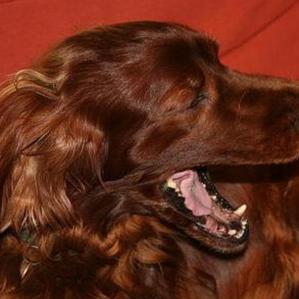}
	\includegraphics[width=0.13\textwidth]{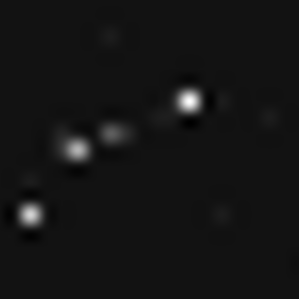}
	\includegraphics[width=0.13\textwidth]{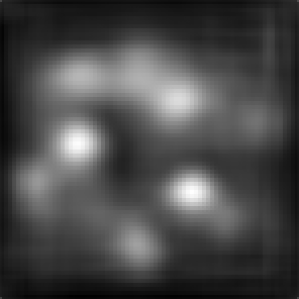}
	\includegraphics[width=0.13\textwidth]{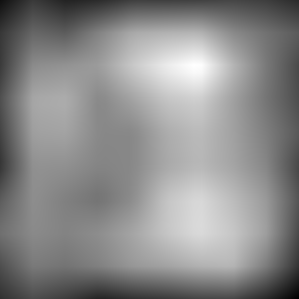}
	\includegraphics[width=0.13\textwidth]{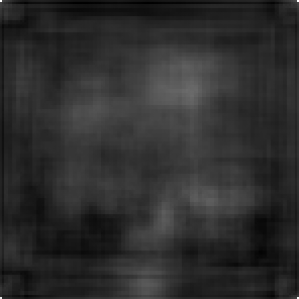}
	\includegraphics[width=0.13\textwidth]{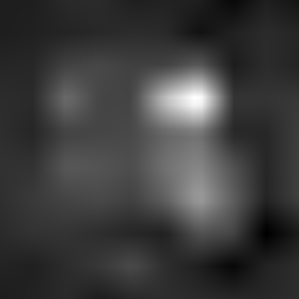}
	\includegraphics[width=0.13\textwidth]{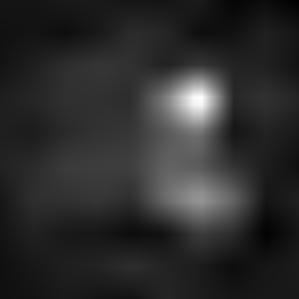}\\ 
	\vspace{0.05cm}
	\includegraphics[width=0.13\textwidth]{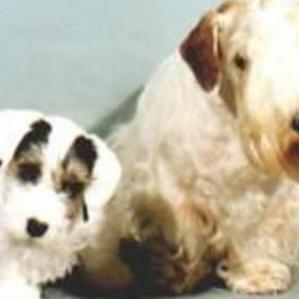}
	\includegraphics[width=0.13\textwidth]{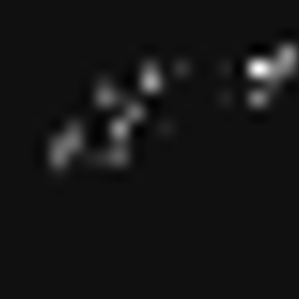}
	\includegraphics[width=0.13\textwidth]{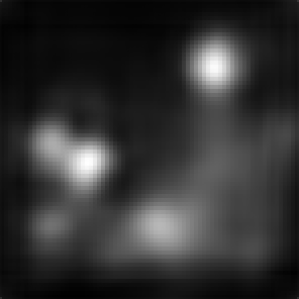}
	\includegraphics[width=0.13\textwidth]{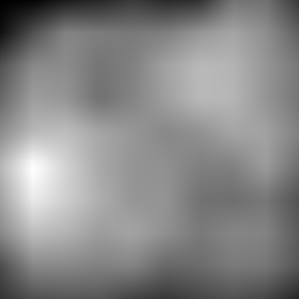}
	\includegraphics[width=0.13\textwidth]{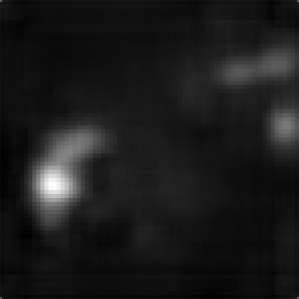}
	\includegraphics[width=0.13\textwidth]{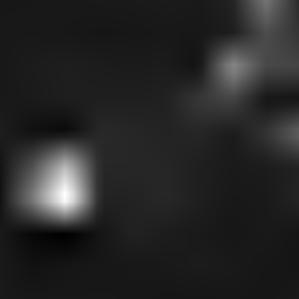}
	\includegraphics[width=0.13\textwidth]{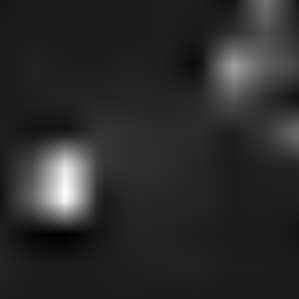}\\ 
	\vspace{0.05cm}
	\includegraphics[width=0.13\textwidth]{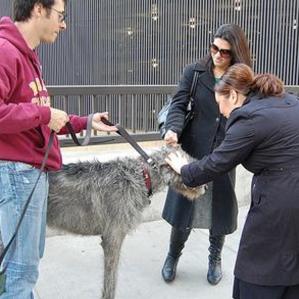}
	\includegraphics[width=0.13\textwidth]{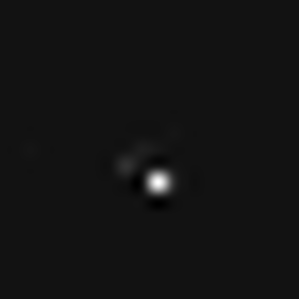}
	\includegraphics[width=0.13\textwidth]{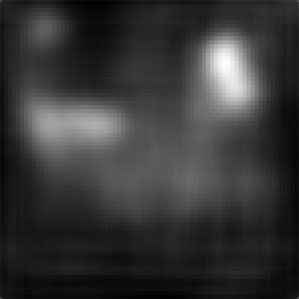}
	\includegraphics[width=0.13\textwidth]{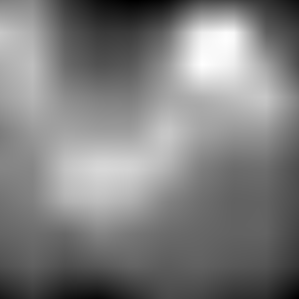}
	\includegraphics[width=0.13\textwidth]{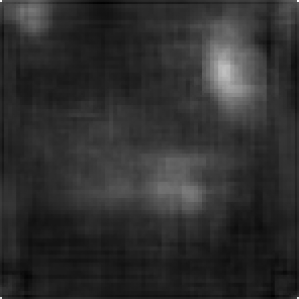}
	\includegraphics[width=0.13\textwidth]{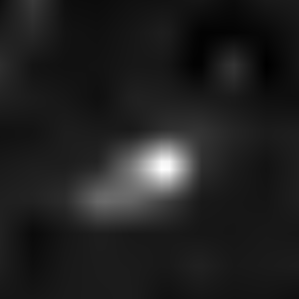}
	\includegraphics[width=0.13\textwidth]{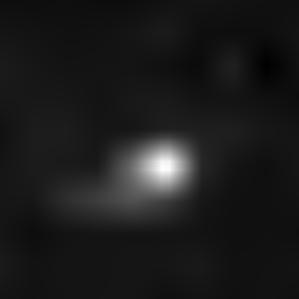}
	\caption{\textbf{Comparison of saliency output maps of different methods}. Each row, from left to right, shows an example image, the corresponding ground-truth saliency map, and the output maps computed, in order, by SalNet and SALICON, as released, and fine-tuned over \emph{SalDogs-train} and the proposed end-to-end SalClassNet model. Beside being able to identify those areas which can be useful for recognition (see first three rows), our method can highlight multiple salient objects (both dogs in the fourth row), or suppress those objects which are not salient for the task (see fifth row).}
	\label{fig:saliency_outputs}
\end{figure*}

\begin{figure*}[h!]
	\centering
	\includegraphics[width=1\textwidth]{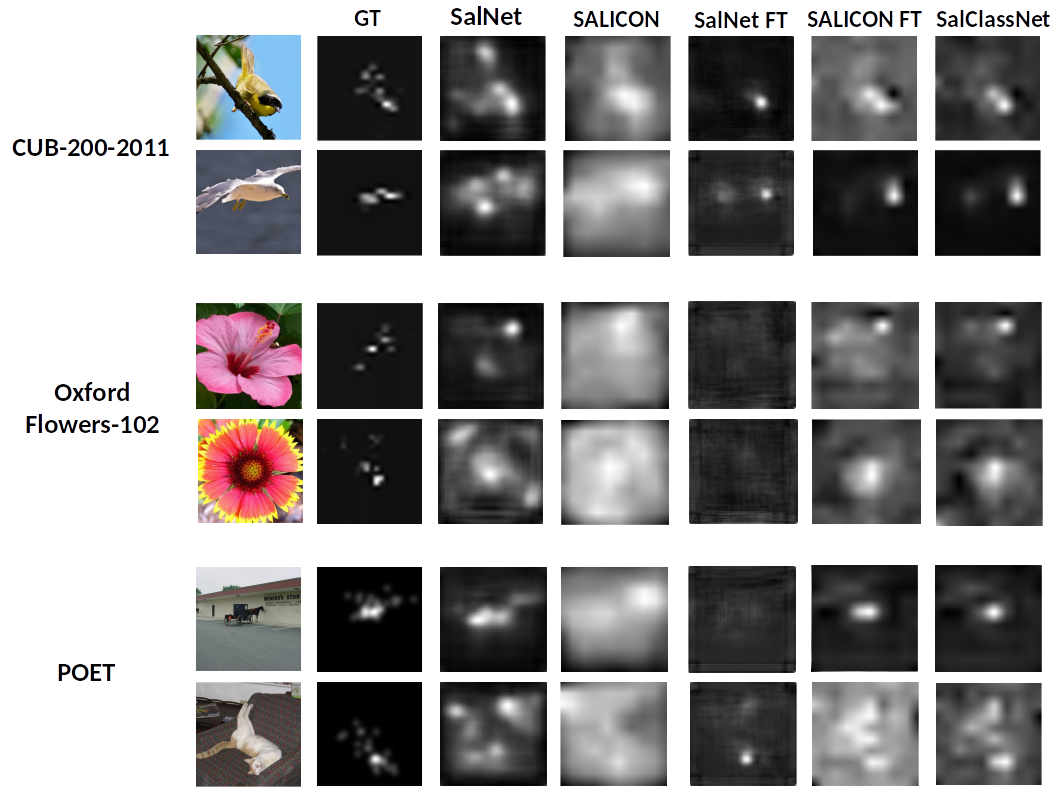}
	\caption{Examples of output saliency maps generated by different methods on \textbf{CUB-200-2011} (first two rows) \textbf{Oxford Flower 102} (third and forth row) and \textbf{POET} (last two rows row) and compared to SALICON-generated saliency maps.
	Each row, from left to right, shows an example image, the corresponding ground-truth saliency map, and the output maps computed, in order, by SalNet and SALICON, both as released and fine-tuned over \emph{SalDogs-train}, and the proposed end-to-end SalClassNet model. SalClassNet, when compared to SALICON (the second best model in Table \ref{tab:results_saliency}), shows better capabilities to filter out image parts which are salient in general but not necessary for classification.}
	\label{fig:saliency_birds_flowers}
\end{figure*}

\vspace{2pt} 
To evaluate the capabilities of SalClassNet for saliency detection, we employed the metrics defined by \cite{borji2013} --- shuffled area under curve (s-AUC), normalized scanpath saliency (NSS) and correlation coefficient (CC) scores --- and compared its performance to those achieved by the SALICON and SalNet models, in their original versions (i.e., as released, pre-trained on the datasets in Tab.~\ref{tab:method_comparison}) and after fine-tuning on \emph{SalDogs-train}.\\
Tab.~\ref{tab:results_saliency} reports a quantitative comparison between these approaches over the \emph{SalDogs-test}, POET, \emph{SalBirds} and \emph{SalFlowers} datasets. It is possible to notice that SalClassNet is able to generate more accurate (and generalizes better) top-down saliency maps than existing methods, which suggests that driving the generation of saliency maps with a specific goal does lead to better performance than fine-tuning already-trained models.
Fig.~\ref{fig:saliency_outputs} and~\ref{fig:saliency_birds_flowers} report some output examples of the tested methods on different input images from, respectively, \emph{SalDogs-}, POET, \emph{SalBirds}, \emph{SalFlowers}. 
Quantitative and qualitative results show SalClassNet's capabilities to generalize well the top-down visual attention process across different datasets. 


\begin{table}\footnotesize
	\centering
	\begin{tabular}{clccc}
		\toprule
		& \textbf{Method} & \textbf{s-AUC} & \textbf{NSS} &  \textbf{CC} \\
		\midrule
		\textbf{Dataset} & \multicolumn{4}{c}{\textbf{SalDogs}}\\
		\midrule
		& Human Baseline    & 0.984&	11.195	&1 \\
		\midrule
		& SalNet    & 0.720&	1.839	&0.231 \\
		& SALICON 	& 0.805	&2.056&	0.261\\
		\midrule
		& Fine-tuned SalNet    	& 0.817	&4.174&	0.432 \\
		& Fine-tuned SALICON 	& 0.837	&3.899&	0.428\\
		& SalClassNet           & 0.862	&4.239&	0.461 \\
		\midrule
		\textbf{Dataset} & \multicolumn{4}{c}{\textbf{POET}}\\
		\midrule
		& Human Baseline    & 0.975&	5.189	&1 \\
		\midrule
		& SalNet    & 0.646	&1.274	&0.342 \\
		& SALICON 	& 0.723	&1.270	&0.355 \\
		\midrule
		& Fine-tuned SalNet     & 0.660	&1.378&	0.300\\
		& Fine-tuned SALICON 	& 0.695	&1.669&	0.356\\
		&SalClassNet            & 0.715	&1.908	&0.387\\
		\midrule
		\textbf{Dataset} & \multicolumn{4}{c}{\textbf{SalBirds}}\\
		\midrule
		& Human Baseline    & 0.743&	9.323	& 1 \\
		\midrule
		& SalNet    & 0.642	&2.252&	0.330  \\
		& SALICON 	& 0.680	&2.247&	0.346  \\

		\midrule
		& Fine-tuned SalNet     & 0.644	&3.504&	0.403 \\
		& Fine-tuned SALICON 	& 0.686	&4.252&	0.507 \\
		&SalClassNet            & 0.708	&4.404&	0.529 \\
		\midrule
		\textbf{Dataset} & \multicolumn{4}{c}{\textbf{SalFlowers}}\\
		\midrule
		& Human Baseline     & 0.975&	9.787	& 1 \\
		\midrule
		& SalNet    & 0.606	&1.311 &0.1973 \\
		& SALICON 	& 0.653	&1.081 &0.1803 \\
		\midrule
		& Fine-tuned SalNet     & 0.576	&0.916	&0.136 \\
		& Fine-tuned SALICON 	& 0.661	&1.599	&0.234 \\
		&SalClassNet            & 0.683	&1.675	&0.245 \\
		\bottomrule
	\end{tabular}
	\caption{Comparison in terms of shuffled area under curve (s-AUC), normalized scanpath saliency (NSS) and correlation coefficient (CC) between the proposed SalClassNet and the baseline models. For each dataset we report the human baseline, i.e., the scores computed using the ground truth maps.}
	\label{tab:results_saliency}
\end{table}

\subsection{Effect of saliency maps on visual classification performance}
\vspace{3pt} 
In this section, we investigate if, and to what extent, explicitly providing saliency maps can contribute to improve classification performance. 
To this end, we first assessed the performance of VGG-19 and Inception over \emph{SalDogs} when using as input a) only color images (3-channel models) and b) ground-truth saliency maps plus color images (4-channel models). In both cases, as mentioned earlier, we re-trained Inception and VGG-19 from scratch, on the de-duped version of ImageNet and then fine-tuned them on \emph{SalDogs-train}, to force the 4-channel versions to use saliency information coming from the upstream module. Indeed, the publicly-available versions of Inception and VGG had already learned dog breed distributions (trained over 150,000 ImageNet dog images), thus they tended to ignore additional inputs such as saliency. Furthermore, a comparison with Inception and VGG-19 pre-trained on the whole ImageNet would have been unfair also because \emph{SalDogs} contains only about 9,000 images (versus 150,000).

We compared the above methods to our SalClassNet, which automatically generates saliency maps and uses them for classification. Besides the version of SalClassNet described in Sect.~\ref{sec:model} (which is also used in all the next experiments), we tested a variant of SalClassNet which employs VGG-19 (suitably modified to account for the saliency input) as classifier: this model is indicated in the results as ``SalClassNet (VGG)''.

Tab.~\ref{tab:results_classification} shows the achieved mean classification accuracies for all the tested methods. It is possible to notice that explicitly providing saliency information (both as ground-truth saliency maps and generated by SalClassNet) to traditional visual classifiers yields improved performance. Indeed, both VGG and Inception suitably extended to make use of saliency information and SalClassNet outperformed the traditional Inception and VGG-19. The lower classification accuracies of the RGBS versions of Inception and VGG (trained with ground truth saliency maps) w.r.t. the SalClassNet variants depend likely by end-to-end training of both saliency and classification networks, which results in extracting and combining, in a more effective way, saliency information with visual cues for the final classification.

\begin{table}
	\centering
	\footnotesize{
	\begin{tabular}{lc}
		\toprule
		\textbf{Method} & \textbf{MCA} \\
		\midrule
		VGG (3 channels)           & 43.4\%\\
		VGG (4 channels) + ground truth saliency maps   & 47.2\%\\
		SalClassNet (VGG)      & 49.0\% \\
		\midrule
		Inception (3 channels)           & 67.1\%\\
		Inception (4 channels) + ground truth saliency maps & 68.4\%\\
		SalClassNet & 70.5\% \\
		\toprule
	\end{tabular}
	\caption{Comparison in terms of mean classification accuracy on \emph{SalDogs-test} between the original Inception and VGG models, pre-trained on \emph{ImageNetDD} (ImageNet without the dog image classes) and fine-tuned on \emph{SalDogs-train}, their RGBS variants trained on ground-truth saliency heatmaps and the respective two variants of SalClassNet.}}
	\label{tab:results_classification}
\end{table}


 Furthermore, SalClassNet showed good generalization capabilities over different datasets, namely, CUB-200-2011 and Oxford Flower 102. In particular, we employed SalClassNet as a feature extractor for a subsequent softmax classifier and compared its performance to those achieved, on the same datasets, by Inception and VGG-19 (fine-tuned on \emph{SalDogs-train} and employed also as feature extractors followed by a softmax classifier). Results are shown in Tab.~\ref{tab:results_features} and confirm our previous claim.
The better generalization performance of our method can be explained by a) the fact that the features learned by the classifiers are not strictly dog-specific, but, more likely, belonging to a wider pattern of fine details that can be generally interpreted as significant features (e.g, eyes, ears, mouth, tail, etc.) for classification, thus applicable to a variety of domains; b) SalClassNet, building and improving on the features by Inception, exploits saliency to weigh better the most distinctive features for classification. Hence, although SalClassNet has not been trained on the flower and bird datasets, the generic nature of the learned features and the improved feature filtering gained through saliency led to high accuracy also on those. In order to demonstrate the effectiveness of SalClassNet’s kernels on different datasets, we computed the features learned by SalClassNet for classification over Stanford Dogs, CUB-200-2011 and Oxford Flowers 102. Table~\ref{tab:visualize_features} shows some of these features, extracted  at different SalClassNet depths and visualized by feeding the whole datasets to the network and identifying the image regions which maximally activate the neurons of certain feature maps. It can be seen how meaningful features for dogs turn out to be meaningful for birds and flowers as well.

\begin{table}
 \centering
 \footnotesize{
 \begin{tabular}{lcc}
 \toprule
 \textbf{} & \textbf{CUB-200-2011} & \textbf{Oxford Flower 102} \\
 \midrule
 \textbf{Method} &  &  \\
 \midrule
 VGG   & 47.6\% & 59.2\% \\
 Inception   & 61.8\% & 77.8\% \\
 SalClassNet & \textbf{63.2}\% & \textbf{79.4}\% \\
 \bottomrule
 \end{tabular}
 \caption{Performance obtained by VGG, Inception and SalClassNet over, respectively, \textbf{CUB-200-2011} and \textbf{Oxford Flower 102}}
 \label{tab:results_features}}
\end{table}

\begin{table*}
 \centering
 \begin{tabular}{lccc}
 \toprule
 \textbf{Layer} & \textbf{Stanford Dogs} & \textbf{CUB-200-2011} & \textbf{Oxford Flower 102} \\
 \midrule
 1 & 
 \begin{minipage}{0.3\textwidth}
  \includegraphics[width=0.2\textwidth]{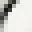} \hspace{5pt}
  \includegraphics[width=0.2\textwidth]{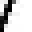}
  \includegraphics[width=0.2\textwidth]{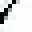}
  \includegraphics[width=0.2\textwidth]{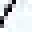} \\
  \includegraphics[width=0.2\textwidth]{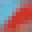} \hspace{5pt}
  \includegraphics[width=0.2\textwidth]{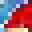}
  \includegraphics[width=0.2\textwidth]{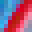}
  \includegraphics[width=0.2\textwidth]{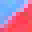} \\
  \includegraphics[width=0.2\textwidth]{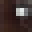} \hspace{5pt}
  \includegraphics[width=0.2\textwidth]{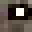}
  \includegraphics[width=0.2\textwidth]{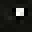}
  \includegraphics[width=0.2\textwidth]{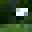}
 \end{minipage} &
 \begin{minipage}{0.3\textwidth}
  \includegraphics[width=0.2\textwidth]{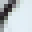} \hspace{5pt}
  \includegraphics[width=0.2\textwidth]{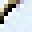}
  \includegraphics[width=0.2\textwidth]{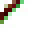}
  \includegraphics[width=0.2\textwidth]{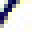} \\
  \includegraphics[width=0.2\textwidth]{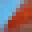} \hspace{5pt}
  \includegraphics[width=0.2\textwidth]{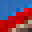}
  \includegraphics[width=0.2\textwidth]{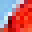}
  \includegraphics[width=0.2\textwidth]{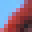} \\
  \includegraphics[width=0.2\textwidth]{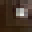} \hspace{5pt}
  \includegraphics[width=0.2\textwidth]{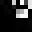}
  \includegraphics[width=0.2\textwidth]{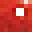}
  \includegraphics[width=0.2\textwidth]{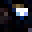}
 \end{minipage} &
 \begin{minipage}{0.3\textwidth}
  \includegraphics[width=0.2\textwidth]{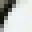} \hspace{5pt}
  \includegraphics[width=0.2\textwidth]{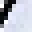}
  \includegraphics[width=0.2\textwidth]{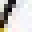}
  \includegraphics[width=0.2\textwidth]{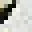} \\
  \includegraphics[width=0.2\textwidth]{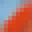} \hspace{5pt}
  \includegraphics[width=0.2\textwidth]{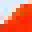}
  \includegraphics[width=0.2\textwidth]{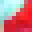}
  \includegraphics[width=0.2\textwidth]{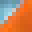} \\
  \includegraphics[width=0.2\textwidth]{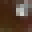} \hspace{5pt}
  \includegraphics[width=0.2\textwidth]{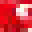}
  \includegraphics[width=0.2\textwidth]{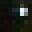}
  \includegraphics[width=0.2\textwidth]{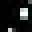}
 \end{minipage} \\
  \midrule
 5 & 
 \begin{minipage}{0.3\textwidth}
  \includegraphics[width=0.2\textwidth]{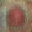} \hspace{5pt}
  \includegraphics[width=0.2\textwidth]{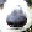}
  \includegraphics[width=0.2\textwidth]{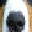}
  \includegraphics[width=0.2\textwidth]{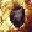} \\
  \includegraphics[width=0.2\textwidth]{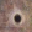} \hspace{5pt}
  \includegraphics[width=0.2\textwidth]{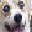}
  \includegraphics[width=0.2\textwidth]{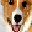}
  \includegraphics[width=0.2\textwidth]{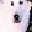} \\
  \includegraphics[width=0.2\textwidth]{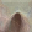} \hspace{5pt}
  \includegraphics[width=0.2\textwidth]{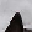}
  \includegraphics[width=0.2\textwidth]{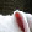}
  \includegraphics[width=0.2\textwidth]{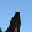}
 \end{minipage} &
 \begin{minipage}{0.3\textwidth}
  \includegraphics[width=0.2\textwidth]{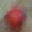} \hspace{5pt}
  \includegraphics[width=0.2\textwidth]{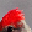}
  \includegraphics[width=0.2\textwidth]{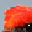}
  \includegraphics[width=0.2\textwidth]{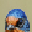} \\
  \includegraphics[width=0.2\textwidth]{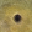} \hspace{5pt}
  \includegraphics[width=0.2\textwidth]{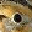}
  \includegraphics[width=0.2\textwidth]{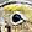}
  \includegraphics[width=0.2\textwidth]{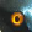} \\
  \includegraphics[width=0.2\textwidth]{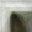} \hspace{5pt}
  \includegraphics[width=0.2\textwidth]{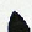}
  \includegraphics[width=0.2\textwidth]{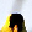}
  \includegraphics[width=0.2\textwidth]{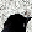}
 \end{minipage} &
 \begin{minipage}{0.3\textwidth}
  \includegraphics[width=0.2\textwidth]{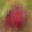} \hspace{5pt}
  \includegraphics[width=0.2\textwidth]{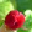}
  \includegraphics[width=0.2\textwidth]{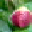}
  \includegraphics[width=0.2\textwidth]{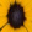} \\
  \includegraphics[width=0.2\textwidth]{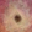} \hspace{5pt}
  \includegraphics[width=0.2\textwidth]{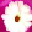}
  \includegraphics[width=0.2\textwidth]{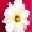}
  \includegraphics[width=0.2\textwidth]{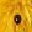} \\
  \includegraphics[width=0.2\textwidth]{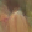} \hspace{5pt}
  \includegraphics[width=0.2\textwidth]{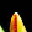}
  \includegraphics[width=0.2\textwidth]{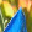}
  \includegraphics[width=0.2\textwidth]{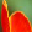}
 \end{minipage} \\
 \midrule
 11 & 
 \begin{minipage}{0.3\textwidth}
  \includegraphics[width=0.2\textwidth]{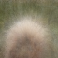} \hspace{5pt}
  \includegraphics[width=0.2\textwidth]{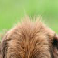}
  \includegraphics[width=0.2\textwidth]{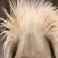}
  \includegraphics[width=0.2\textwidth]{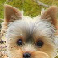} \\
  \includegraphics[width=0.2\textwidth]{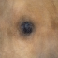} \hspace{5pt}
  \includegraphics[width=0.2\textwidth]{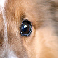}
  \includegraphics[width=0.2\textwidth]{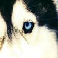}
  \includegraphics[width=0.2\textwidth]{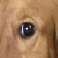} \\
  \includegraphics[width=0.2\textwidth]{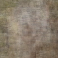} \hspace{5pt}
  \includegraphics[width=0.2\textwidth]{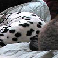}
  \includegraphics[width=0.2\textwidth]{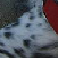}
  \includegraphics[width=0.2\textwidth]{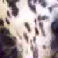}
 \end{minipage} &
 \begin{minipage}{0.3\textwidth}
  \includegraphics[width=0.2\textwidth]{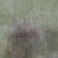} \hspace{5pt}
  \includegraphics[width=0.2\textwidth]{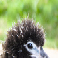}
  \includegraphics[width=0.2\textwidth]{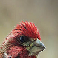}
  \includegraphics[width=0.2\textwidth]{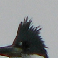} \\
  \includegraphics[width=0.2\textwidth]{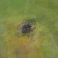} \hspace{5pt}
  \includegraphics[width=0.2\textwidth]{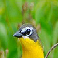}
  \includegraphics[width=0.2\textwidth]{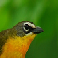}
  \includegraphics[width=0.2\textwidth]{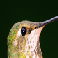} \\
  \includegraphics[width=0.2\textwidth]{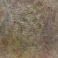} \hspace{5pt}
  \includegraphics[width=0.2\textwidth]{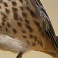}
  \includegraphics[width=0.2\textwidth]{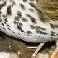}
  \includegraphics[width=0.2\textwidth]{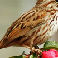}
 \end{minipage} &
 \begin{minipage}{0.3\textwidth}
  \includegraphics[width=0.2\textwidth]{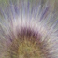} \hspace{5pt}
  \includegraphics[width=0.2\textwidth]{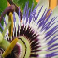}
  \includegraphics[width=0.2\textwidth]{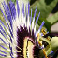}
  \includegraphics[width=0.2\textwidth]{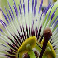} \\
  \includegraphics[width=0.2\textwidth]{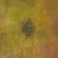} \hspace{5pt}
  \includegraphics[width=0.2\textwidth]{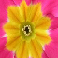}
  \includegraphics[width=0.2\textwidth]{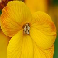}
  \includegraphics[width=0.2\textwidth]{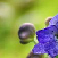} \\
  \includegraphics[width=0.2\textwidth]{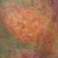} \hspace{5pt}
  \includegraphics[width=0.2\textwidth]{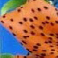}
  \includegraphics[width=0.2\textwidth]{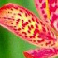}
  \includegraphics[width=0.2\textwidth]{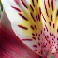}
 \end{minipage} \\
 \midrule
 16 & 
 \begin{minipage}{0.3\textwidth}
  \includegraphics[width=0.2\textwidth]{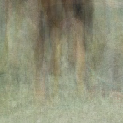} \hspace{5pt}
  \includegraphics[width=0.2\textwidth]{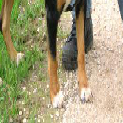}
  \includegraphics[width=0.2\textwidth]{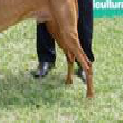}
  \includegraphics[width=0.2\textwidth]{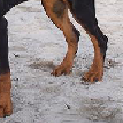} \\
  \includegraphics[width=0.2\textwidth]{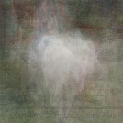} \hspace{5pt}
  \includegraphics[width=0.2\textwidth]{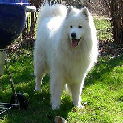}
  \includegraphics[width=0.2\textwidth]{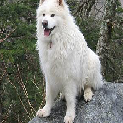}
  \includegraphics[width=0.2\textwidth]{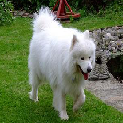} \\
  \includegraphics[width=0.2\textwidth]{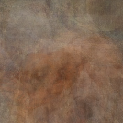} \hspace{5pt}
  \includegraphics[width=0.2\textwidth]{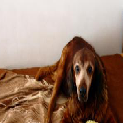}
  \includegraphics[width=0.2\textwidth]{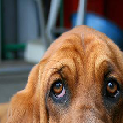}
  \includegraphics[width=0.2\textwidth]{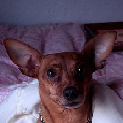}
 \end{minipage} &
 \begin{minipage}{0.3\textwidth}
  \includegraphics[width=0.2\textwidth]{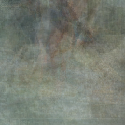} \hspace{5pt}
  \includegraphics[width=0.2\textwidth]{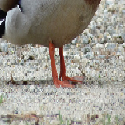}
  \includegraphics[width=0.2\textwidth]{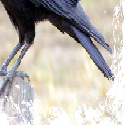}
  \includegraphics[width=0.2\textwidth]{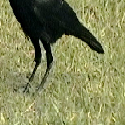} \\
  \includegraphics[width=0.2\textwidth]{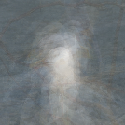} \hspace{5pt}
  \includegraphics[width=0.2\textwidth]{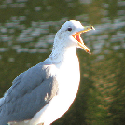}
  \includegraphics[width=0.2\textwidth]{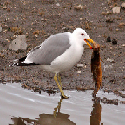}
  \includegraphics[width=0.2\textwidth]{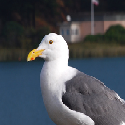} \\
  \includegraphics[width=0.2\textwidth]{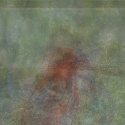} \hspace{5pt}
  \includegraphics[width=0.2\textwidth]{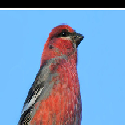}
  \includegraphics[width=0.2\textwidth]{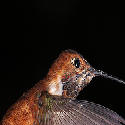}
  \includegraphics[width=0.2\textwidth]{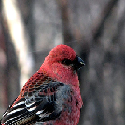}
 \end{minipage} &
 \begin{minipage}{0.3\textwidth}
  \includegraphics[width=0.2\textwidth]{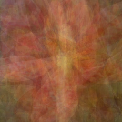} \hspace{5pt}
  \includegraphics[width=0.2\textwidth]{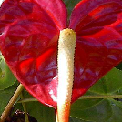}
  \includegraphics[width=0.2\textwidth]{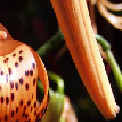}
  \includegraphics[width=0.2\textwidth]{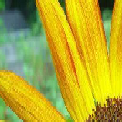} \\
  \includegraphics[width=0.2\textwidth]{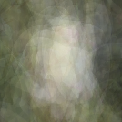} \hspace{5pt}
  \includegraphics[width=0.2\textwidth]{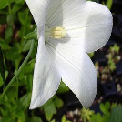}
  \includegraphics[width=0.2\textwidth]{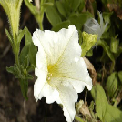}
  \includegraphics[width=0.2\textwidth]{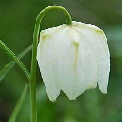} \\
  \includegraphics[width=0.2\textwidth]{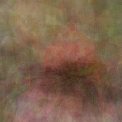} \hspace{5pt}
  \includegraphics[width=0.2\textwidth]{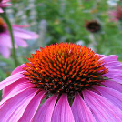}
  \includegraphics[width=0.2\textwidth]{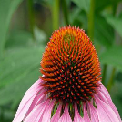}
  \includegraphics[width=0.2\textwidth]{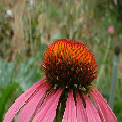}
 \end{minipage} \\
 \bottomrule
 \end{tabular}
 \caption{Examples of features employed by SalClassNet for classification over three different datasets. Each row of images in the tables shows sample which provide high activations for a certain feature map. For each of the tested datasets, we show a 3$\times$4 block of images, where the first column represents the average image computed over the highest 50 activations for that dataset; the last three columns show the three top activations.}
 \label{tab:visualize_features}
\end{table*}
\section{Concluding remarks}
In this work, we proposed a deep architecture --- SalClassNet --- which generates top-down saliency maps by conditioning, through the object class supervision, the saliency detection process and, at the same time, exploits such saliency maps for visual classification. Performance analysis, both in terms of saliency detection and classification, showed that SalClassNet identifies regions corresponding to class-discriminative features, hence emulating top-down saliency, unlike most of the existing saliency detection methods which produce bottom-up maps of generic salient visual features.
Although we tested our framework using two specific networks for saliency detection and visual classification, 
its architecture and our software implementation are general and can be used with any fully-convolutional saliency detector or classification network by simply replacing one of the two subnetworks, respectively, before or after the connecting batch normalization module.
As further contribution of this paper, we built a dataset of saliency maps (by means of eye-gaze tracking experiments on 12 subjects who were asked to guess dog breeds) for a subset of the Stanford Dog dataset, creating what is, to the best of our knowledge, the first publicly-available top-down saliency dataset driven by a fine-grained visual classification task.
We hope that our flexible deep network architecture (all source code is available) together with our eye-gaze dataset will push the research in the direction of emulating human visual processing through a deeper understanding of the higher-level (such as top-down visual attention) processes behind it. 
\vspace{20pt}

\bibliographystyle{model2-names}
\bibliography{egbib}

\end{document}